\documentclass[runningheads]{llncs}

\usepackage{eccv}

\usepackage{eccvabbrv}

\usepackage{graphicx}
\usepackage{booktabs}

\usepackage[accsupp]{axessibility}  

\usepackage{microtype}
\usepackage{enumitem}

\usepackage{hyperref}

\usepackage{orcidlink}
\usepackage[ruled]{algorithm2e}
\SetKwInOut{Input}{Input}
\SetKwInOut{Output}{Output}
\begin{document}

\sloppy

\title{LNL+K: Enhancing Learning with Noisy Labels Through Noise Source Knowledge Integration} 

\titlerunning{LNL+K}

\author{Siqi Wang\orcidlink{0009-0007-6861-3806} \and
Bryan A. Plummer\orcidlink{0000-0002-7074-3219} }

\authorrunning{S. Wang et al.}

\institute{Boston University, Boston MA 02215, USA\\
\email{\{siqiwang,bplum\}@bu.edu}}

\maketitle

\begin{abstract}
Learning with noisy labels (LNL) aims to train a high-performing model using a noisy dataset. 
We observe that noise for a given class often comes from a limited set of categories, yet many LNL methods overlook this.
For example, an image mislabeled as a cheetah is more likely a leopard than a hippopotamus due to its visual similarity.  Thus, we explore Learning with Noisy Labels with noise source Knowledge integration (LNL+K), which leverages knowledge about likely source(s) of label noise that is often provided in a dataset's meta-data.   Integrating noise source knowledge boosts performance even in settings where LNL methods typically fail.  For example, LNL+K methods are effective on datasets where noise represents the majority of samples, which breaks a critical premise of most methods developed for LNL.  Our LNL+K methods can boost performance even when noise sources are estimated rather than extracted from meta-data.
We provide several baseline LNL+K methods that integrate noise source knowledge into state-of-the-art LNL models that are evaluated across six diverse datasets and two types of noise, where we report gains of up to 23\% compared to the unadapted methods. Critically, we show that LNL methods fail to generalize on some real-world datasets, even when adapted to integrate noise source knowledge, highlighting the importance of directly exploring LNL+K\footnote{Code available: \textbf{\url{https://github.com/SunnySiqi/LNL_K}}}. 
\end{abstract}
\section{Introduction}


\begin{figure}[t]
  \centering
    \includegraphics[width=\textwidth]{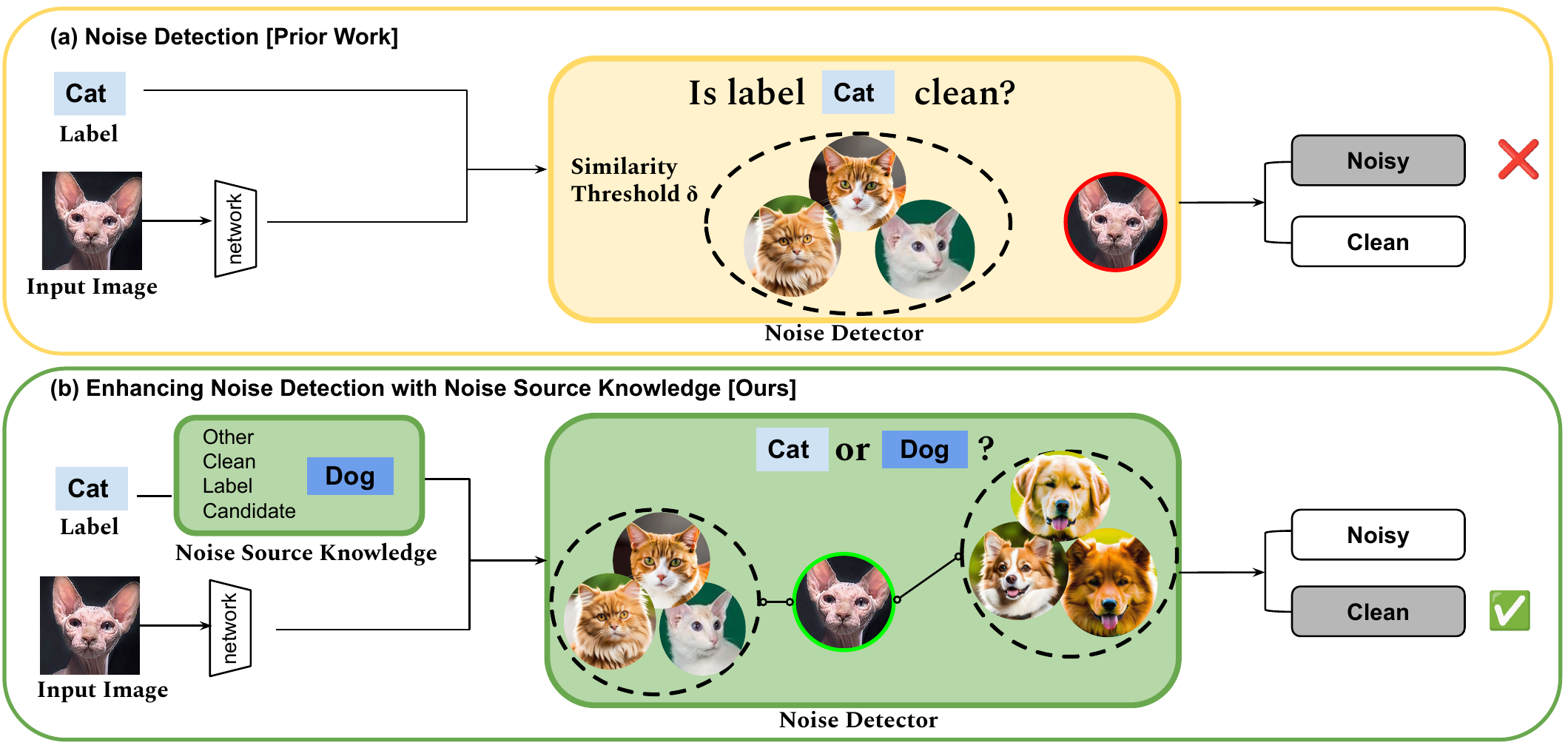}
  \caption{(Best view in color.) \textbf{Comparison of LNL and LNL+K on a hard-negative clean sample}.
(a) Traditional LNL methods (\eg,~\cite{kim2021fine, crust, wei2022self, karim2022unicon}) classify an input image as having a noisy label based on a similarity threshold between the sample and its (majority) class features. (b) In contrast, LNL+K methods identify the sample as a clean label by considering the noise source \textit{dog}. Specifically, since probability of \textit{cat} is higher than that of \textit{dog} and aligns more closely with the \textit{cat} in the feature space, LNL+K judges it as more likely a \textit{cat} image.  
}
  \label{fig:fig1}
\end{figure}

High-quality labeled data is valuable for training deep neural networks, but it's costly and often corrupted in real-world datasets~\cite{krishna2016embracing, yan2014learning}. Learning with Noisy Labels (LNL)~\cite{natarajan2013LNL}  aims to learn from noisy training data while achieving strong generalization performance~\cite{arpit2017closer, song2022survey}. 
Prior work addresses this task along two main themes: one aligns the noisy data classifier with the clean data classifier through estimated noise transitions~\cite{scott2015rate, liu2015classification, yao2020dual, xia2019anchor, zhang2021learning, kye2022learning, cheng2022instance}, while the other discriminates between noisy and clean samples~\cite{kim2021fine, crust, wei2022self, liu2020ELR,iscen2022ncr, han2018co-teaching,karim2022unicon, li2020dividemix}.
The core challenge in both types of methods centers on distinguishing potential clean and noisy samples.
For example, in Fig.~\ref{fig:fig1}-a the input image contains a cat that looks dissimilar to the other cat samples. Thus, prior work would find it challenging to identify this sample as clean (\eg~\cite{kim2021fine, crust, wei2022self, karim2022unicon}). However, as shown in Fig.~\ref{fig:fig1}-b, with external knowledge, such as knowing that if the \textit{cat} is mislabeled, it is more likely to be mislabeled as a \textit{dog}, we can identify this sample as a clean label.


We observe that noise source knowledge already exists or can be estimated in real-world datasets. Labels are rarely uniformly corrupted across all classes, and some classes are more easily confused than others~\cite {tanno2019estimation_annotator}. 
\Eg, visually similar objects are often mislabeled: wolf and coyote~\cite{song2019selfie}, automobiles and trucks~\cite{krizhevsky2009cifar}. Furthermore, in scientific settings, certain categories are intentionally designed to establish causality and can be treated as noise sources during training.  These are often referred to as a \textit{control} \ie, \textit{do-nothing} group~\cite{control-wiki}. For example, cell painting images are labeled with a treatment applied to the cells, but those treatments may have little-to-no effect, meaning that most cells would visually resemble the \textit{control} class (\ie, their true label should be \textit{control}). The noise ratio in this setting can be over 50\%~\cite{rohban2017systematic}, which means prior work~\cite{kim2021fine, crust, wei2022self, karim2022unicon} would be prone to incorrectly consider the more prominent noisy images as the \textit{true} label.
Thus, integrating noise source knowledge offers significant potential, particularly in scientific domains with high noise ratios. 

To this end, we explore \textbf{L}earning with \textbf{N}oisy \textbf{L}abels through noise source \textbf{K}nowledge integration (LNL+K). In contrast to traditional LNL tasks, we assume that we are given some knowledge about noisy label distribution. \ie, noisy labels tend to originate from specific categories (\eg, that the dog is the potential noise source Fig.~\ref{fig:fig1}-b). The integration of knowledge about noise sources helps discriminate clean samples in two ways. First, it aids in the identification of hard negative instances.  Second, it enables us to detect noise even at high noise ratios.  These two benefits come arise from a shared cause: that our goal is not to identify what labels are clean, but rather what labels are more likely from a noise source.
To illustrate, in Fig.~\ref{fig:fig1}-b the probability that the input image is a cat is low, but it more like the cat instances than the dogs (the noise source), so it would still be recognized as a clean label. When there are many noisy images, resulting in all cat images producing low probability. LNL+K would separate those that also have high probability of being from a noise source (as noisy) from images with predictions that are higher for non-noise source categories (as clean).



Han~\etal~\cite{han2018masking} is the most similar work to ours, which introduces a form of noise supervision by removing invalid noise transitions through human cognition.  Thus, while Han~\etal also aims to leverage some knowledge about the noise source, they focus on estimating noise transitions to avoid overfitting to noisy labels. 
In our paper, we update and greatly expand on their initial work, including introducing a unified framework with which we can adapt recent methods from LNL to our LNL+K task (\eg,~\cite{crust, kim2021fine, wei2022self, karim2022unicon, li2023disc}) 
and investigating new noise settings designed to reflect applications to scientific datasets. 
Our experiments report up to an 8\% gain under asymmetric noise and a remarkable 15\% accuracy boost under dominant noise on synthesized noise using CIFAR-10/CIFAR-100~\cite{krizhevsky2009cifar}. We also obtain a 1-2\% gain on a diverse set of four real-world noisy datasets including two image-based cell profiling datasets~\cite{pratapa2021image}, CHAMMI-CP~\cite{CHAMMI} and BBBC036~\cite{bray2016cell}, and two natural image datasets, Animal-10N~\cite{song2019selfie} and Clothing1M~\cite{xiao2015learning}.

In summary, our contributions are:
\begin{itemize}[nosep,leftmargin=*]
  \item  We explore an important but overlooked task, termed \textbf{LNL+K}: Enhancing \textbf{L}earning with \textbf{N}oisy \textbf{L}abels through noise source \textbf{K}nowledge integration. We also design a new noise setting that is widespread in real world: dominant noise, where noisy samples can be the majority of a labeled category distribution.
  \item We define a unified framework for clean label detection in LNL+K that we use to adapt LNL with noise source knowledge to serve as baselines for LNL+K. 
  \item We analyze the robustness of LNL+K methods on incomplete and noisy knowledge, explore estimating noise source knowledge from noise transition estimation methods, and define \textit{knowledge absorption rate} which can measure how well an LNL method can be transferred to LNL+K tasks, providing an optimizing objective for improving LNL+K methods.
\end{itemize}
\section{Related Work}
\noindent\textbf{\textit{LNL with noise supervision} methods} are precursors to the broader concept of LNL+K~\cite{hendrycks2018using, li2017learning, yu2023delving, veit2017learning, han2018masking}. \textit{Noise supervision} refers to possessing prior knowledge about the noise present in the dataset. For example, a small clean dataset provides noise source supervision to create more accurate estimates of the noise distribution. However, obtaining human-verified clean datasets is costly and often unavailable. Han~\etal~\cite{han2018masking} propose using human cognition of invalid class transitions as \textit{mask} to reduce the burden of transition matrix estimation. However, this approach is constrained to classifier-consistent methods, and the outcomes become unreliable when the noise structure is misidentified. 
In contrast to these methods, our LNL+K task does not require complete knowledge.  
Our goal is to use any existing knowledge about noise sources in a dataset, and we find even partial or noisy noise source knowledge can boost performance.  
\smallskip

\noindent\textbf{\textit{Classifier-consistent} methods} align a classifier trained on noisy data with the optimal classifier, which often minimizes errors on clean data. Given input $X$ with true label $Y$ and noisy label $\widetilde{Y}$ the goal is to infer the clean class posterior probability $P(Y|X = x)$ using the noisy class posterior probability $P(\widetilde{Y}|X = x)$ (which can be learned using noisy data) and the transition matrix $T(X = x)$ where $T_{ij}(X = x) = P(\widetilde{Y}= j|Y = i, X = x)$.  To estimate the transition matrix, some work uses \textit{anchor points}, samples with very high probability of belonging to a certain class~\cite{scott2015rate, liu2015classification, menon2015learning, patrini2017making}. To avoid using additional clean data, some work focuses on estimating the transition matrix with noisy data~\cite{li2021provably, yao2020dual, xia2019anchor, zhang2021learning, kye2022learning, cheng2022instance, liu2023identifiability, li2022estimating}. This can be achieved with the density ratio estimation method~\cite{vapnik2013constructive} and matrix decomposition~\cite{yao2020dual, patrini2017making, liu2015classification}, or by training a network to predict the transition matrix~\cite{yong2022holistic, yang2022estimating}. 
Statistically consistent methods train both noisy and clean data indiscriminately but heavily depend on the accuracy of the noise transition matrix, which becomes particularly challenging with high noise ratios. As we will show, LNL+K can indirectly enhance datasets that require noise source estimation by combining our proposed task with methods for estimating the noise source.
\smallskip

\noindent\textbf{\textit{Classifier-inconsistent} methods} discriminate between clean and noisy labels and handle them differently during training. To discriminate between clean and noisy samples many methods use losses that detect noisy samples with high loss values~\cite{jiang2018mentornet, li2020dividemix, arazo2019unsupervised_loss}, and some probability-distribution-based approaches select clean samples with high confidence~\cite{hu2021p, torkzadehmahani2022confidence, nguyen2019self, tanaka2018joint, li2022selective, feng2021ssr}. However, these assumptions may not always hold true, especially with hard negative samples along distribution boundaries. \Ie, samples selected by these approaches are more likely \textit{easy} samples instead of \textit{clean} samples. 
Feature-based approaches utilize the input before the softmax layer -- high-dimensional features~\cite{crust, kim2021fine}, which are less affected by noisy labels~\cite{li2020lastlayeroverfitting, yao2020searching, bai2021understanding}. 
To differentiate the training of noisy and clean samples, there are methods adjusting the loss function~\cite{wei2022self,ma2020normalized,iscen2022ncr,xu2019l_dmi, zhang2018generalized}, using regularization techniques~\cite{liu2020ELR,xia2021param, hu2019simple}, multi-round learning only with selected clean samples~\cite{cordeiro2023longremix,shen2019learning_iterative, wu2020topological}, and training noisy samples with semi-supervised learning (SSL) techniques~\cite{sohn2020fixmatch, tarvainen2017mean, li2020dividemix, karim2022unicon, li2023disc}. 
To our knowledge, most statistically inconsistent methods often overlook the valuable resource of noise distribution knowledge in the context of LNL. LNL+K makes a unique contribution by utilizing noise source knowledge to detect clean samples.

\section{Learning with Noisy Labels + Knowledge (LNL+K)}
\label{sec:task_definition}
Learning with Noisy Label Source Knowledge (LNL+K) aims to find the parameter set $\theta^*$ for the classifier $f_\theta$ that achieves the highest accuracy on the clean test set when trained on the noisy dataset $D$ \textbf{with noise source knowledge $D_{ns}$}. 
Suppose we have a dataset $D = \{(x_i, \widetilde{y_i})_{i=1}^n \in R^d \times K\}$, where $K = \{1, 2, ..., k\}$ is the categorical label for $k$ classes. $(x_i, \widetilde{y_i})$ denotes the $i-th$ example in the dataset, such that $x_i$ is a $d-dimiensional$ input in $R^d$ and $\widetilde{y_i}$ is the label. $\{\widetilde{y_i}\}_{i=1}^n$ might include noisy labels and the true labels $\{y_i\}_{i=1}^n$ are unknown. However, we have some prior knowledge about noisy label sources. 
This knowledge can take various forms—it can be precise, such as the noise transition matrix obtained through noise modeling methods, or it can be imprecise and incomplete, stemming from human cognition, \eg, classes like \textit{cat} and \textit{lynx} are visually similar and are more likely mislabeled with each other~\cite{song2019selfie}.  Additionally, knowledge can be derived from the dataset design, \eg \textit{control} class serves as the noise source in scientific datasets.
Thus, the noise source distribution knowledge $D_{ns}$ can be represented in different ways. One representation is by a probability matrix  $P_{k \times k}$, where $P_{ij}$ refers to the probability that a sample in class $i$ is mislabeled as class $j$. Alternatively, it can also be represented using a set of label pairs  $LP = \{(i, j) | i, j \in K\}$, where $(i, j)$ indicates that samples in class $i$ are more likely to be mislabeled as class $j$. 
For the convenience of formulating the following equations, noise source knowledge $D_{c-ns}$ represents the set of noise source labels of category $c$. \Ie, $D_{c-ns} = \{i | i\in K \land (P_{ic}>0 \lor (i,c) \in LP)\}$. 

\subsection{A Unified Framework for Clean Sample Detection with LNL+K}
\label{subsec:framework}

To make our framework general enough to represent different LNL methods, we define a unified logic of clean sample detection. 
Formally, consider sample $x_i$ with a clean categorical label $c$, \ie, 
\begin{equation}
y_i =c \leftrightarrow \widetilde{y_i} = c \land p(c|x_i) > \delta,
\label{eq:1}
\end{equation} where $p(c|x_i)$ is the probability of sample $x_i$ with label $c$ and $\delta$ is the threshold for the decision. Different methods vary in how they obtain $p(c|x_i)$.  For example, 
loss-based detection uses $Loss(f_\theta(x_i), c)$ to estimate $p(c|x_i)$~\cite{jiang2018mentornet, li2020dividemix, arazo2019unsupervised_loss}, probability-distribution-based methods use the logits or classification probability score $f_\theta(x_i)$~\cite{hu2021p, torkzadehmahani2022confidence, nguyen2019self, tanaka2018joint, li2022selective}, and feature-based methods use $p(c|x_i)=M(x_i,\phi_c)$~\cite{crust, kim2021fine}, where $M$ is a similarity function, $\phi_c = D(g(X_c))$ is the distribution of features labeled as category $c$, \ie, $X_c = \{x_i | \widetilde{y_i}=c\}$, $g(X_c) = \{g(x_i, c) | x_i \in X_c\} \sim \phi_c$, and $g(\cdot)$ is a feature mapping function. Feature-based methods often vary in their feature mapping $g(\cdot)$ function and similarity function $M$. 

LNL+K adds knowledge $D_{ns}$ by comparing $p(c|x_i)$ with $p(c_n|x_i)$, where $c_n$ is the noise source label. When category $c$ has multiple noise source labels, $p(c|x_i)$ should be greater than any of them. In other words, the probability that sample $x_i$ has label $c$ (\ie, $p(c|x_i)$), not only depends on its own value, but how it compares to noise sources. \Eg, the cat input image in Fig.~\ref{fig:fig1} has low predicted probability of being a cat, \ie, $p(cat|x_i) < \delta$, so LNL methods would identify it as noise (shown in Fig.~\ref{fig:fig1}-a). However, for LNL+K in Fig.~\ref{fig:fig1}-b, we compare the likelihood of this image being a cat to the probability of belonging to the noise source dog class. Thus, since $p(cat|x_i) > p(dog|x_i)$, it is marked a clean sample. 

To summarize, the propositional logic of LNL+K is: 
\begin{equation}
y_i  =c \leftrightarrow \widetilde{y_i}= c \land p(c|x_i) > \max(\{p(c_n|x_i) | c_n \in D_{c-ns}\}).
\label{eq:2}
\end{equation}
 Whereas conventional LNL methods would select examples with the highest probably for a given class, Eq.~\ref{eq:2} in LNL+K differs in that:
\begin{enumerate}[nosep,leftmargin=*]
\item The selected sample's probability may not be its highest predicted category. This is particularly beneficial for identifying hard negative samples, such as images with similar backgrounds. \Ie, while the objects themselves may not exhibit similar features and are not considered noise sources, the shared background can cause model confusion for LNL methods. 
\item We introduce cross-class comparisons with feature similarity. Most existing LNL methods utilizing feature space similarity do not incorporate cross-class comparisons, focusing solely on the likelihood samples belong to their designated class (Eq.~\ref{eq:1}). Although AUM~\cite{pleiss2020identifying} introduced cross-class comparisons to the non-assigned label with the highest probability, it was limited to logits and performed poorly at high noise levels like those we study in this paper.
\end{enumerate}
    

\subsection{Incorporating Noise Source Knowledge into LNL methods}
\label{sec:baselines}

We adapt several recent methods from the LNL literature as baselines for our LNL+K task. 
Our adaptations enhance the detection of clean samples in \textit{inconsistent-classifier methods}. Once the probability of samples being clean is determined, the remainder of the training process follows the original methods. We provide a summary of each adaptation below, but additional details can be found in the supplementary. 
Algorithm~\ref{alg:input_knowledge_algo} summarizes the steps of our framework, offering a unified approach to integrating noise source knowledge through cross-class comparisons. Note that each base model primarily differs in the function $p(c|x_i)$, which calculates the probability of a sample being clean.
\smallskip


\noindent\textbf{CRUST$^{+k}$} adapts CRUST~\cite{crust}, which uses the pairwise gradient distance within the class for clean sample detection. A clean sample subset is selected with the most similar gradients clustered together.  
To estimate the likelihood of a sample label being clean in CRUST$^{+k}$, 
we apply CRUST on the combined sample set of label class and noise source class. Specifically, for a target label class $c$, for each noise source $c_{ns}$ in $D_{c-ns}$, we create the union set of samples $\{x_i |\widetilde{y_i} = c \lor \widetilde{y_i} = c_{ns}\}$.  Then CRUST is applied on this union set to identify the clean samples for noise source class $c_{ns}$. If a sample with label $c$ is selected as part of the clean cluster of $c_{ns}$, we assume its label is noisy. Let $CRUST(x, c) > 0$ be the indicator that sample $x$ is identified as a clean sample for label $c$ through the computation of the gradient directed towards label $c$, where $x$ falls within the cluster of similar gradients. Thus,  clean $c$ labeled samples identified by CRUST$^{+k}$ are those that satisfy the following: $\{x | CRUST(x, c_{ns}) <= 0, \forall c_{ns} \in D_{c-ns}\}$.
\smallskip

\noindent\textbf{FINE$^{+k}$} is derived from FINE~\cite{kim2021fine}, utilizing the alignment between sample and label class features for detection. This alignment is determined by the cosine distance between the sample's features and the eigenvector of the class feature gram matrix, serving as the feature representation for that category. FINE employs a Gaussian Mixture Model (GMM) on the alignment distribution to categorize samples into clean and noisy groups.
FINE$^{+k}$ enhances this by incorporating noise source class information into the alignment calculation. In FINE$^{+k}$, the clean probability is the difference between the alignment with the label class and the noise-source-class alignment. If $g(x)$ represents the sample feature and $G(c)$ is the feature representation of class $c$, then FINE fits a GMM directly on $Sim<g(x), G(c)>$ similarity values, while FINE$^{+k}$ fits a GMM on the alignment difference scores between the labeled class and noise source classes. \Ie, $Sim<g(x), G(c)> - \max(\{Sim<g(x), G(n_{ns})>| n_{ns} \in  D_{c-ns}\})$. 
\smallskip

\noindent\textbf{SFT$^{+k}$} is based on SFT~\cite{wei2022self}, which identifies noisy samples by comparing their predictions over a few recent epochs. A sample is detected as noisy if it used to be classified correctly, but it is misclassified in the latest epoch. SFT$^{+k}$ is adapted by restricting the misclassified labels only to noise source labels. 
\smallskip

\noindent\textbf{UNICON$^{+k}$} is adapted from UNICON~\cite{karim2022unicon}, which estimates the clean probability by using Jensen-Shannon divergence (JSD), a metric for distribution dissimilarity. Disagreement between predicted and one-hot label distributions is utilized, ranging from 0 to 1, with smaller values indicating a higher probability of the label being clean. UNICON$^{+k}$ integrates the noise source knowledge by adding the comparison of JSD with the noise source class. If the sample's predicted distribution aligns more with the noise source, it is considered noisy. For sample $x$ with label $\widetilde{y}$ and noise source knowledge about $\widetilde{y}$ as $D_{\widetilde{y}-ns}$, the clean samples detected by UNICON$^{+k}$ are those belonging to $\{x | JSD(x, \widetilde{y}) < \min(\{JSD(x, c_n) | c_n \in D_{y-ns}\})\}$.
\smallskip

\noindent\textbf{DISC$^{+k}$} adapts DISC~\cite{li2023disc}, which identifies clean and noisy samples based on predictions from two diverse augmentations. Clean samples are selected if the confidences of predictions on weak and strong augmentation images both are over certain thresholds. The clean set is defined as $\{x,\widetilde{y} |p_w(\widetilde{y}, x) > \tau_w (t)\} \cap \{x,\widetilde{y} |p_s(\widetilde{y}, x) > \tau_s (t)\}$, where $p(\widetilde{y}, x)$ represents prediction confidences (\textit{w}: weak, \textit{s}: strong), and $\tau (t)$ is the dynamic instance-specific threshold (DIST). The formula for $\tau (t)$ is given by $\tau_x (t) = \lambda \tau (t-1) + (1-\lambda) p_x (t)$, where $p_x (t) = \max(\{p(c; x) | c \in K\})$. In DISC$^{+k}$, the adaptation involves assigning values to $p_x (t)$ by selecting the largest value among the label class and noise source classes, i.e., $p_{x}(t) = \max(\{p(c; x) | c \in D_{\widetilde{y}-ns}\} \cup \{p(\widetilde{y}; x)\})$.
\smallskip

\noindent\textbf{DualT+X$^{+k}$} combines noise estimation and noise discrimination methods.  
DualT~\cite{yao2020dual} is a consistent-classifier method that estimates noise transitions by factorizing the transition matrix into two new matrices that are often easier to estimate compared to the original matrix. Its estimated noise transition matrix can serve as input for any LNL+K method denoted as X$^{+k}$. 
We use a straightforward approach to obtain the noise sources from the noise transition matrix: for each class $c$, we select the class $c_ns$ with the second-highest transition probability (as the highest probability corresponds to the class itself) as the noise source.

\begin{algorithm}[t]
\caption{Noise Source Integration Algorithm.}\label{alg:input_knowledge_algo}
\Input{Inputs $X=\{x_i\}_{i=1}^n$, noisy labels $Y=\{\widetilde{y_i}\}_{i=1}^n$, probability function $p$ in adaptation-base-method, noise source knowledge $D_{ns}$}
\Output{Probabilities of samples being clean $P(X) = \{p(\widetilde{y_i}|x_i)\}_{i=1}^n$}
\For{$i\gets1$ \KwTo $n$}{ 
    $p_i \gets p(\widetilde{y_i}|x_i)$ \tcp*[l]{Probability of given label $\widetilde{y_i}$ being clean.}
    \For{c in $D_{ns}$}{ \tcp{Loop through noise sources.}
        \If{$p(c|x_i) > p_i$}{
        \tcc{If $x_i$ is more likely to belong to the noise source label $c$, then $\widetilde{y_i}$ is considered as the noisy label.}
        $p_i \gets 0$\; 
        break\;
        }
    }
}
\end{algorithm}


\section{Experiments}
We benchmark two types of noise across six diverse datasets on LNL+K. Each type of noise is evaluated using both synthesized noise from CIFAR-10/CIFAR-100~\cite{krizhevsky2009cifar} datasets and two real-world noisy datasets. CIFAR-10/CIFAR-100~\cite{krizhevsky2009cifar} dataset contains 10/100 classes with 5,000/500 images per class for training and 1,000/100 images per class for testing, respectively. Synthesized noisy labels are generated based on the noise type for training and validation sets, while ground truth labels are retained for testing and result analysis. For real-world noisy datasets, noise source knowledge is found in the dataset's meta-data (\eg example confusion matrix~\cite{xiao2015learning} or experiment design of \textit{controls} as a noise source for cell datasets~\cite{bray2016cell}), \ie, no new annotations are obtained.  
Additional experimental setup and implementation details can be found in the supplementary.
\smallskip

\noindent\textbf{Baselines.} 
In addition to the baseline methods detailed in Section~\ref{sec:baselines}, we introduce three additional points of comparison. First, \textit{Baseline} trains on noisy datasets without modifications, \ie, it employs no LNL or LNL+K methods. Second, we include two \textit{noise transition estimation methods}: DualT~\cite{yao2020dual} and GT-T, where GT-T denotes the method trained with a ground truth transition matrix, serving as an upper bound for methods focused on estimating this matrix (\eg, DualT~\cite{yao2020dual}). Note GT-T only applies to synthesized noise experiments due to the absence of ground truth in real-world datasets. The ground truth transition matrix is also used as the noise source knowledge by our $+k$ methods on CIFAR. Third, we compare to SOP~\cite{pmlr-v162-liu22w}, a regularization-focused LNL approach.

\subsection{Dominant Noise Experiments}

\begin{figure}[tbh]
\centering
\begin{minipage}[c]{0.5\textwidth}
\includegraphics[width=\textwidth]{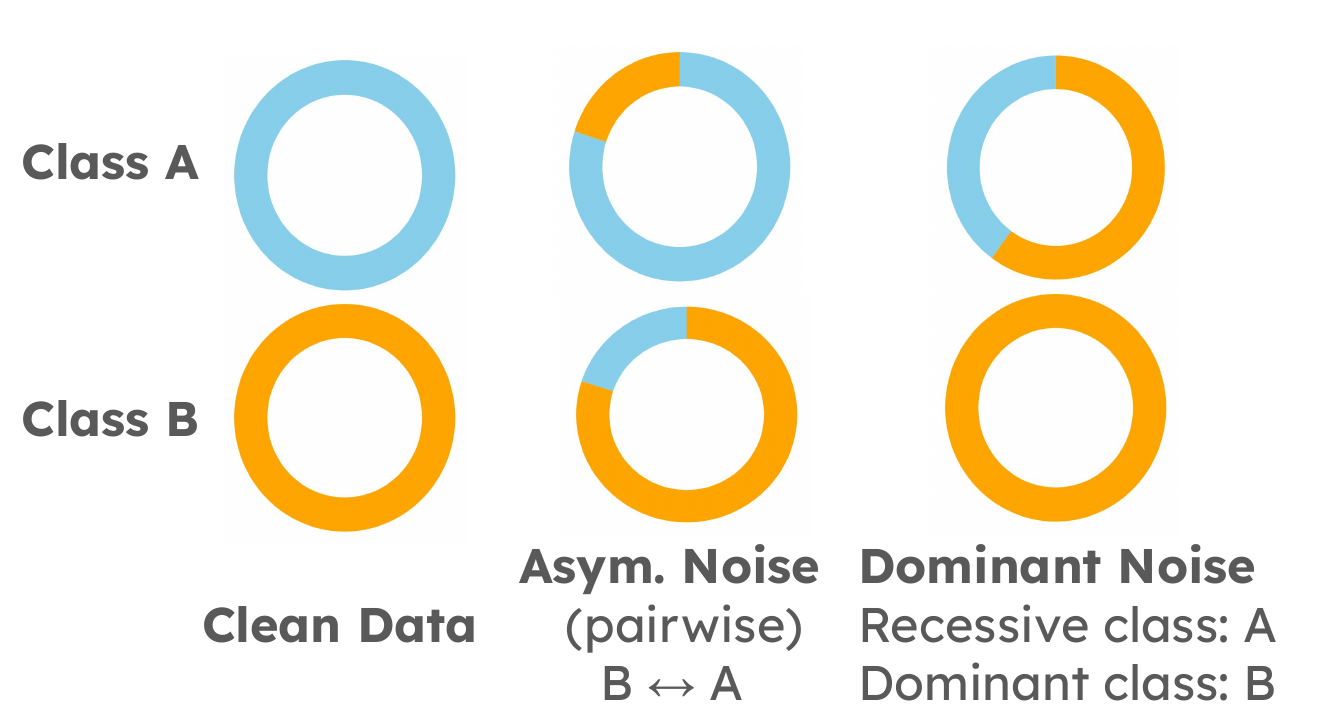} 
\end{minipage}%
\hspace{0.5cm}
\begin{minipage}[c]{0.4\textwidth}
\caption{\textbf{Comparisons of noise types.} Asymmetric noise can occur bidirectionally with limited noise ratios, while Dominant noise can exceed 50\% in recessive classes, with clean dominant classes.}
\label{fig:dominant_noise}
\end{minipage}
\end{figure}

\label{subsec:dominant}
\noindent\textbf{Synthesized Dominant Noise and Noisy Cell Image Datasets}
\begin{itemize}[nosep,leftmargin=*]
\item \textbf{Dominant Noise} is a novel setting designed to simulate high-noise ratios in real-world settings, particularly in scientific datasets. In this setup, we categorize classes as either \textit{dominant} or \textit{recessive}, where samples mislabeled as \textit{recessive} likely belong to the \textit{dominant} class. \Eg in Fig. \ref{fig:dominant_noise}, the noisy samples of \textit{recessive} class A are from  \textit{dominant} class B. Our dominant noise simulation can be seen as a special case of unbalanced asymmetric noise with three key features mirroring noise in scientific datasets: 1) The noise ratio can exceed 50\%, reflecting scenarios where experiments fail to produce a significant effect, resulting in a high noise ratio~\cite{rohban2017systematic}. 2) The noise source is always known, as scientists need a \textit{baseline} for their experiments. 3) The noise is one-directional, with the \textit{control} class intentionally devoid of noise sources. 

For CIFAR-10/CIFAR-100~\cite{krizhevsky2009cifar}, we define half the categories as \textit{recessive} and the other half as \textit{dominant}. Noisy labels are generated by labeling images in \textit{dominant} classes as \textit{recessive}, thus \textit{dominant} classes act as noise sources for the \textit{recessive} classes. \textit{Dominant} noise can create a skewed distribution (see example in Fig.~\ref{fig:dominant_noise}), which challenges the informative dataset assumption used by prior work~\cite{cheng-instance}. To maintain balance after label corruption, we select different sample numbers from \textit{recessive} and \textit{dominant} classes, see supplementary for detailed noise composition information.

\item\textbf{Cell Datasets BBBC036 and CHAMMI-CP} 
contain single U2OS cell (human bone osteosarcoma) images from Cell Painting datasets~\cite{bray2016cell}, which represent large treatment screens of chemical and genetic perturbations. Each treatment is tested and then imaged with the Cell Painting protocol~\cite{bray2016cell}, which is based on six fluorescent markers captured in five channels. 
Our goal is to classify the effects of treatments with cell morphology features trained by the model. A significant challenge is that cells react differently to the treatment, \ie,  some show minimal differences from controls, creating noisy labels where images look like \textit{controls} (\textit{doing-nothing} group) but are labeled as treatments. 
Approximately 1,300 of the 1,500 treatments show high similarity with the \textit{controls}~\cite{bray2016cell}, suggesting \textbf{the majority of the weak-treatment cell images may be noisy}.
BBBC036\footnote{Available at \textbf{\url{https://bbbc.broadinstitute.org/image\_sets}}} and CHAMMI-CP\footnote{Available at \textbf{\url{https://zenodo.org/record/7988357}}} sampled single-cell images from 1,500 bioactive compounds (treatments), including the \textit{control} group.
For BBBC036, we sampled 100 treatments, one of which is labeled as \textit{controls}, resulting in 132,900 training, 14,257 validation, and 22,016 test images. For CHAMMI-CP, we removed three treatments that only appeared in the test set, resulting in four classes, \textit{weak},  \textit{medium}, \textit{strong} treatments, and \textit{control}, with 36,360 training, 3,636 validation, and 13,065 test images. The \textit{control} category is considered the noise sources for all other classes. 
\end{itemize}
\smallskip

\begin{table*}[t]
  \centering
  \caption{Results for dominant noise on CIFAR-10 and CIFAR-100~\cite{krizhevsky2009cifar} datasets, along with Cell datasets~\cite{CHAMMI, bray2016cell}. The best test accuracy is marked in bold, and the better result between LNL and LNL+K methods are underlined. We find incorporating source knowledge helps in most cases. See Section~\ref{subsec:dominant} for discussion.}
    \begin{tabular}{lcccccc}
    \toprule
     & \multicolumn{2}{c}{CIFAR-10~\cite{krizhevsky2009cifar}} & \multicolumn{2}{c}{CIFAR-100~\cite{krizhevsky2009cifar}} & CHAMMI-CP & BBBC036 \\
     \cmidrule(lr){2-3}  \cmidrule(lr){4-5}
     Noise ratio  & 0.5 & 0.8 & 0.5 & 0.8 & \cite{CHAMMI} & \cite{bray2016cell} \\
    \midrule
    Baseline  & 85.46\scriptsize{$\pm$0.25} & 78.99\scriptsize{$\pm$0.07}  & 41.41\scriptsize{$\pm$1.47} & 27.03\scriptsize{$\pm$0.12} & 71.54\scriptsize{$\pm$0.45} & 63.49\scriptsize{$\pm$0.62}\\
    DualT~\cite{yao2020dual} & 83.70\scriptsize{$\pm$0.04}  & 46.96\scriptsize{$\pm$0.07}  & 27.04\scriptsize{$\pm$0.07} & 19.94\scriptsize{$\pm$0.04} & 70.73\scriptsize{$\pm$0.17} & 61.54\scriptsize{$\pm$0.61} \\
    GT-T  & 85.24\scriptsize{$\pm$0.06} & 76.03\scriptsize{$\pm$0.04}  & 48.39\scriptsize{$\pm$0.21} & 35.96\scriptsize{$\pm$0.04} & - & - \\ 
    SOP~\cite{pmlr-v162-liu22w}  & 86.94\scriptsize{$\pm$0.37} & 80.65\scriptsize{$\pm$0.71}  & 55.78\scriptsize{$\pm$0.68} & 45.94\scriptsize{$\pm$0.62} & 77.55\scriptsize{$\pm$0.23} & 60.94\scriptsize{$\pm$0.38} \\
    \midrule
    CRUST~\cite{crust}  & 80.46\scriptsize{$\pm$0.17} & 65.79\scriptsize{$\pm$0.62}  & 48.87\scriptsize{$\pm$0.31} & 35.56\scriptsize{$\pm$1.38} & 78.02\scriptsize{$\pm$0.31} & 63.06\scriptsize{$\pm$0.65} \\
    CRUST$^{+k}$   & \underline{87.19\scriptsize{$\pm$0.08}} & \underline{80.54\scriptsize{$\pm$0.30}} & \underline{51.56\scriptsize{$\pm$0.31}} &\underline{38.07\scriptsize{$\pm$2.05}} &\underline{\textbf{79.81}\scriptsize{$\pm$0.56}} & \underline{\textbf{65.07}\scriptsize{$\pm$0.71}}\\
    \midrule
    FINE~\cite{kim2021fine}  & 84.43\scriptsize{$\pm$0.38} &75.45\scriptsize{$\pm$0.74}  & 52.87\scriptsize{$\pm$0.98} & 39.45\scriptsize{$\pm$0.25} & \underline{67.27\scriptsize{$\pm$0.82}} & 56.80\scriptsize{$\pm$0.87}\\
    FINE$^{+k}$  & \underline{88.00\scriptsize{$\pm$0.11}} & \underline{80.52\scriptsize{$\pm$0.28}}  & \underline{54.77\scriptsize{$\pm$1.68}} & \underline{42.25\scriptsize{$\pm$0.27}} & 67.02\scriptsize{$\pm$0.73}  & \underline{57.01\scriptsize{$\pm$0.40}}\\
    \midrule
    SFT~\cite{wei2022self}  & 85.43\scriptsize{$\pm$0.13} & 75.43\scriptsize{$\pm$0.12}  & 48.21\scriptsize{$\pm$1.21} & \underline{41.76\scriptsize{$\pm$1.34}} & 76.08\scriptsize{$\pm$0.25} & 51.71\scriptsize{$\pm$0.82}\\
    SFT$^{+k}$  & \underline{87.31\scriptsize{$\pm$0.15}} & \underline{76.78\scriptsize{$\pm$0.38}}  & \underline{51.21\scriptsize{$\pm$1.14}} & 37.96\scriptsize{$\pm$0.05} & \underline{77.75\scriptsize{$\pm$0.42}} & \underline{59.18\scriptsize{$\pm$1.33}} \\
    \midrule
    UNICON~\cite{karim2022unicon}  & 88.43\scriptsize{$\pm$0.14} & 81.37\scriptsize{$\pm$0.43}   & 57.92\scriptsize{$\pm$0.43}  & 42.70\scriptsize{$\pm$0.50} & \underline{71.45\scriptsize{$\pm$0.03}} & 33.98\scriptsize{$\pm$1.03} \\
    UNICON$^{+k}$  & \underline{89.21\scriptsize{$\pm$0.42}}  & \underline{82.27\scriptsize{$\pm$0.29}}  & \underline{61.55\scriptsize{$\pm$0.13}}  & \underline{48.47\scriptsize{$\pm$0.40}} & 71.04\scriptsize{$\pm$0.14} & \underline{42.17\scriptsize{$\pm$0.31}}\\
    \midrule
    DISC~\cite{li2023disc}  & 91.58\scriptsize{$\pm0.21$} & 85.89\scriptsize{$\pm0.16$} & 64.97\scriptsize{$\pm0.17$} & 49.79\scriptsize{$\pm0.20$} & 74.04\scriptsize{$\pm0.11$} & 40.55\scriptsize{$\pm0.18$} \\
    DISC$^{+k}$  & \underline{\textbf{91.88}\scriptsize{$\pm$0.15}} & \underline{\textbf{86.70}\scriptsize{$\pm$0.03}} & \underline{\textbf{65.96}\scriptsize{$\pm$0.15}} & \underline{\textbf{50.74}\scriptsize{$\pm$0.11}} & \underline{75.38\scriptsize{$\pm0.30$}} & \underline{63.32\scriptsize{$\pm0.49$}}\\
    \bottomrule
  \end{tabular}
  \label{tb:dominant}
\end{table*}

\noindent\textbf{Results.} Table \ref{tb:dominant} reports classification accuracy in dominant noise setting, where  LNL+K methods report the best performance in all cases.  We see that as the noise ratio is increased, the gains reported by LNL+K methods also increase. 
In the synthetic noise scenarios, there is an average improvement of 3\% in the 80\% noise ratio, with CRUST$^{+k}$ achieving an impressive 15\% improvement on CIFAR-10. For the cell datasets, high feature similarity between certain treatments and the \textit{control} group can lead to significantly high noise ratios, strongly influencing the class distribution. 
On BBBC036, DISC$^{+k}$ achieves a significant 23\% boost, but this only brings the method in line with standard training.  Notably, only CRUST$^{+k}$ outperforms standard training, boosting top-1 accuracy by 1.5\%, whereas all LNL methods fail to improve performance.  This helps illustrate the importance of exploring LNL+K.
We also note that in both cell datasets there are additional sources of noise that are not addressed by LNL+K.  Specifically, there are technical variations that arise due to differences in the experimental protocol used to collect the images~\cite{pratapa2021image}.  Thus, our noise source knowledge should be considered incomplete, yet still provide enough signal to boost performance.

To understand the source of our performance gains, in Table~\ref{tb:sample} we report the effect integrating knowledge in LNL+K has on identifying clean samples.  We find that our LNL+K boosts sample selection accuracy across all methods, with the largest gain being an improvement in recall by over 60\% for CRUST while also improving precision by 15\%.  Due to this remarkable gain in sample selection accuracy, CRUST$^{+k}$ performs on par or better than FINE$^{+k}$ and SFT$^{+k}$ on cell datasets, eliminating the apparent advantage these methods had over CRUST in the LNL setting. 
To show the effect that knowledge has on noisy class predictions during training in cell datasets, Fig.~\ref{fig:CP_heatmap} reports the predictions made on weak treatment cells (\ie, the class with the highest noise) on CHAMMI-CP's test set during training with CRUST.  We find that including knowledge enables our approach to more effectively identify these treatments. 
\smallskip

\begin{table}[t]
\begin{minipage}{0.48\textwidth}
\centering
     \captionof{table}{Results for 0.8 dominant noise on CIFAR-10~\cite{krizhevsky2009cifar}. We report precision and recall for the selected \textit{clean} subset and the model's classification accuracy. Integrating knowledge (+K) significantly enhances sample selection performance. See Section~\ref{subsec:dominant} for further discussion.}
      \begin{tabular}{lccc}
     \toprule
       & Precision & Recall & Cls.\ Acc.\ \\
      \midrule
    CRUST~\cite{crust} & 72.29 & 36.15 & 65.79 \\
    CRUST$^{+k}$ & \textbf{87.67} & \textbf{99.04} & \textbf{80.54} \\
    \midrule
    FINE~\cite{kim2021fine} & 88.53 & 61.89 & 75.45 \\
    FINE$^{+k}$ & \textbf{89.64} & \textbf{99.61} & \textbf{80.52} \\
    \midrule
    SFT~\cite{wei2022self} & 97.27 & 94.37 & 75.43 \\
    SFT$^{+k}$ & \textbf{98.99} & \textbf{94.95} & \textbf{76.78} \\
     \bottomrule
     \end{tabular}
     \label{tb:sample}
\end{minipage}\hfill
\begin{minipage}{0.48\textwidth}
    \centering
    \includegraphics[width=\textwidth]{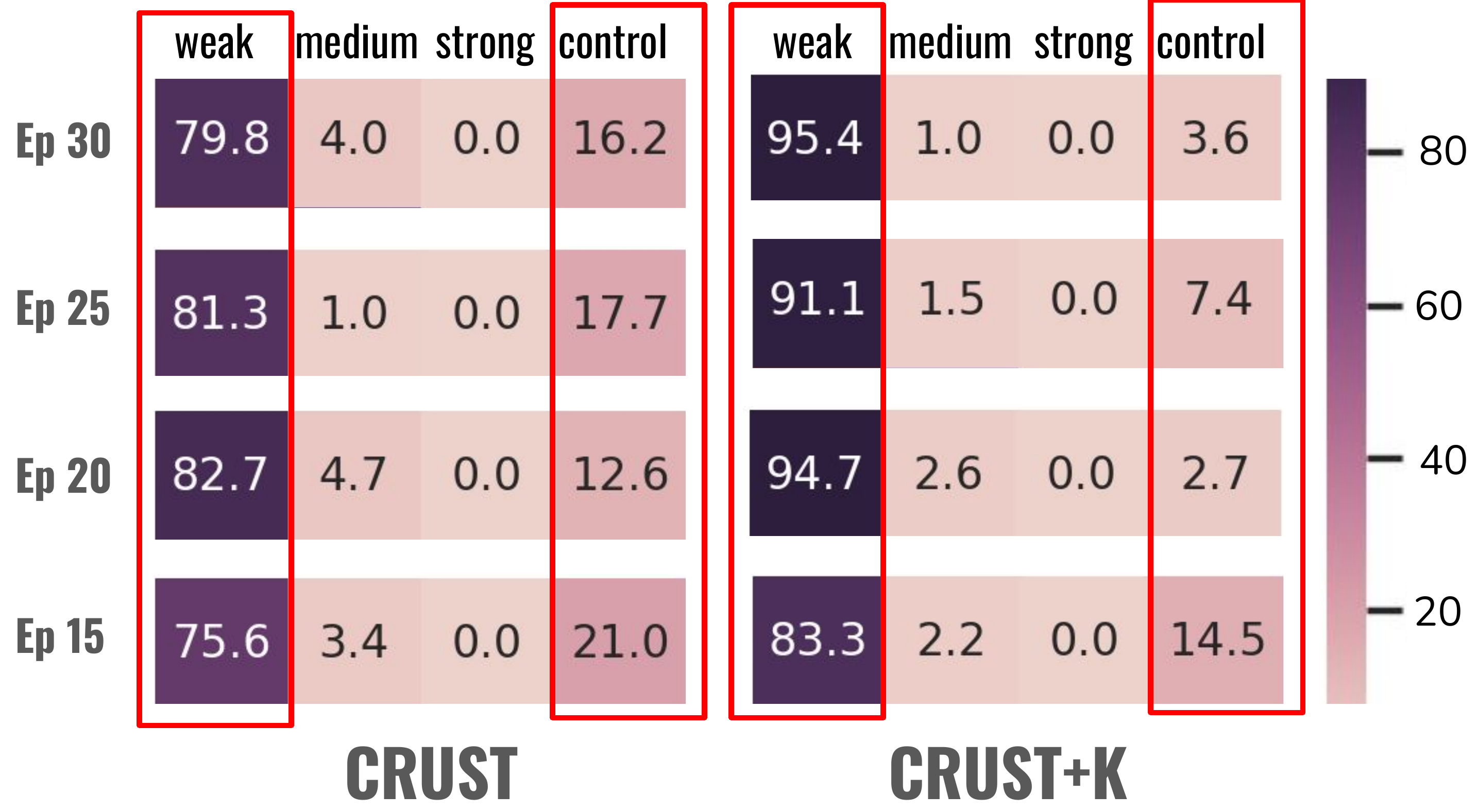}
\captionof{figure}{Class prediction confusion matrix for \textbf{weak treatment} cell images in the CHAMMI-CP~\cite{CHAMMI} dataset, normalized to sum to 100\%. The integration of knowledge (+K) enhances the method's capability to distinguish weak treatment from a high-ratio \textit{control} class. See Section \ref{subsec:dominant} for more details.}
 \label{fig:CP_heatmap}
\end{minipage}
\end{table}

\subsection{Asymmetric Noise Experiments}
\label{subsec:asym}
\noindent\textbf{Synthesized Asymmetric Noise and Noisy Online Image Datasets}
\begin{itemize}[nosep,leftmargin=*]
\item \textbf{Asymmetric Noise} simulates real-world scenarios where visually similar objects are mislabeled as each other. \Ie, labels are corrupted for visually similar classes, such as trucks $\leftrightarrow$ automobiles in CIFAR-10~\cite{krizhevsky2009cifar}. Asymmetric noise is synthesized in a \textbf{pair-wise} manner, \ie, $P(A \to B) = P(B \to A)$. As this noise is bidirectional, when the noise ratio surpasses 50\%, the model lacks cues to discern the true label from the minority class, therefore, our experiments focus on two noise ratios: 20\% and 40\%. The supplementary contains a list of confusing pairs, which serve as noise sources in our experiments.

\item \textbf{Online Image Datasets Animal-10N and Clothing1M} contain images that were obtained by crawling several online search engines using predefined labels as search keywords or extracting noisy labels from surrounding text. The noisy labeled data is utilized for training and validation, while a smaller test set has manually annotated clean labels. 
Animall-10N~\cite{song2019selfie} has 10 confusing animal classes with a total of 50,000 training images and 5000 testing images. The dataset author noted 5 pairs of classes that can be easily confused\footnote{Available at \textbf{\url{https://dm.kaist.ac.kr/datasets/animal-10n/}}}, which serve as noise sources in our experiments. 
Clothing1M~\cite{xiao2015learning} contains approximately one million clothing images.  The dataset encompasses 14 classes, with an estimated overall label accuracy of around 60\%. However, the label noise is imbalanced, with certain classes experiencing noise levels as high as 80\% (\eg, mislabeling \textit{sweater} as \textit{knitwear}). We provide a summary of the noise sources for each class in the supplementary, extracted from the confusion matrix presented in the original paper~\cite{xiao2015learning}. Additional details on the integrated noise source knowledge in our experiments is also in the supplementary.
\end{itemize}

\begin{table}[t]
\caption{Results for asymmetric noise on CIFAR-10 and CIFAR-100~\cite{krizhevsky2009cifar} datasets, along with Animal-10N~\cite{song2019selfie}. The best test accuracy is marked in bold, and the better result between LNL and LNL+K methods are underlined. We find knowledge-adapted methods can alter the rankings of the base methods. (\eg SFT and FINE at a noise ratio of 0.4 on CIFAR-100.) See Section~\ref{subsec:asym} for discussion. }
  \centering
  \setlength{\tabcolsep}{1.5pt}
  \begin{tabular}{lccccc}
    \toprule
     & \multicolumn{2}{c}{CIFAR-10~\cite{krizhevsky2009cifar}} & \multicolumn{2}{c}{CIFAR-100~\cite{krizhevsky2009cifar}} & Animal-10N \\
     \cmidrule(lr){2-3} \cmidrule(lr){4-5} 
     Noise ratio & \multicolumn{1}{c}{0.2} & \multicolumn{1}{c}{0.4} & \multicolumn{1}{c}{0.2} & \multicolumn{1}{c}{0.4} & \cite{song2019selfie} \\
    \midrule
    Baseline & 86.12\scriptsize{$\pm$0.42} & 77.18\scriptsize{$\pm$0.30} & 62.96\scriptsize{$\pm$0.12}  & 59.07\scriptsize{$\pm$0.08} & 80.32\scriptsize{$\pm$0.20} \\
    DualT~\cite{yao2020dual} & 92.24\scriptsize{$\pm$0.10} & 66.23\scriptsize{$\pm$0.03} & 53.61\scriptsize{$\pm$1.49} &52.03\scriptsize{$\pm$1.92} & 81.14\scriptsize{$\pm$0.28} \\
    GT-T & 92.51\scriptsize{$\pm$0.03} & 89.68\scriptsize{$\pm$0.13} & 73.88\scriptsize{$\pm$0.04} & 66.61\scriptsize{$\pm$0.03} & - \\
    SOP~\cite{pmlr-v162-liu22w} & 92.85\scriptsize{$\pm$0.49} & 89.93\scriptsize{$\pm$0.25} & 72.60\scriptsize{$\pm$0.70} & 70.58\scriptsize{$\pm$0.30} & 83.93\scriptsize{$\pm$0.35}\\
    \midrule
    CRUST~\cite{crust} & \underline{91.94\scriptsize{$\pm$0.05}} & \underline{89.40\scriptsize{$\pm$0.03}} & 60.75\scriptsize{$\pm$1.87} & 59.79\scriptsize{$\pm$0.89} & \underline{81.88\scriptsize{$\pm$0.13}}\\
    CRUST$^{+k}$ & 89.47\scriptsize{$\pm$0.17} & 84.96\scriptsize{$\pm$0.91}& \underline{62.44\scriptsize{$\pm$0.84}} & \underline{61.07\scriptsize{$\pm$0.16}} & 81.74\scriptsize{$\pm$0.08}\\
    \midrule
    FINE~\cite{kim2021fine} & 89.07\scriptsize{$\pm$0.03} & 85.51\scriptsize{$\pm$0.18} & 65.42\scriptsize{$\pm$0.11}  & 65.11\scriptsize{$\pm$0.11}& 81.15\scriptsize{$\pm$0.11}\\
    FINE$^{+k}$ & \underline{90.87\scriptsize{$\pm$0.04}} & \underline{89.15\scriptsize{$\pm$0.26}} & \underline{73.59\scriptsize{$\pm$0.12}} & \underline{72.87\scriptsize{$\pm$0.11}} & \underline{82.27\scriptsize{$\pm$0.10}}\\
    \midrule
    SFT~\cite{wei2022self} & 92.67\scriptsize{$\pm$0.04} & 89.77\scriptsize{$\pm$0.14} & \underline{74.41\scriptsize{$\pm$0.05}} & 69.51\scriptsize{$\pm$0.06} & 82.24\scriptsize{$\pm$0.10}\\
    SFT$^{+k}$ & \underline{93.19\scriptsize{$\pm$0.08}} & \underline{90.55\scriptsize{$\pm$0.06}}& 74.29\scriptsize{$\pm$0.14} & \underline{70.94\scriptsize{$\pm$0.13}} & \underline{82.88\scriptsize{$\pm$0.18}}\\
    \midrule
    UNICON~\cite{karim2022unicon} & 92.42\scriptsize{$\pm$0.04} & \underline{91.51\scriptsize{$\pm$0.12}} & 75.95\scriptsize{$\pm$0.04} & 73.08\scriptsize{$\pm$0.07} & 87.76\scriptsize{$\pm$0.06}\\
    UNICON$^{+k}$ & \underline{92.60\scriptsize{$\pm$0.07}} & 91.35\scriptsize{$\pm$0.24}& \underline{76.87\scriptsize{$\pm$0.24}} & \underline{73.97\scriptsize{$\pm$0.11}} & \underline{\textbf{88.28}\scriptsize{$\pm$0.29}} \\
    \midrule
    DISC~\cite{li2023disc} & 94.82\scriptsize{$\pm$0.04} & 93.24\scriptsize{$\pm$0.04} & 76.02\scriptsize{$\pm$0.15} & 74.36\scriptsize{$\pm$0.16} & \underline{86.44\scriptsize{$\pm0.14$}}\\
    DISC$^{+k}$ & \underline{\textbf{95.40}\scriptsize{$\pm$0.08}}  & \underline{\textbf{94.05}\scriptsize{$\pm$0.07}} & \underline{\textbf{77.13}\scriptsize{$\pm$0.05}} & \underline{\textbf{75.50}\scriptsize{$\pm$0.08}}  & 86.90\scriptsize{$\pm0.10$}\\
    \bottomrule
  \end{tabular}
  \label{tb:asym}
\end{table}

\begin{table}[t]
\caption{Results on Clothing1M~\cite{xiao2015learning} dataset. Results with * are from the referenced paper, others are our implementation. See Section~\ref{subsec:asym} for discussion.}
  \centering
  \begin{tabular}{ccccc|cc|cc}
    \toprule
    Baseline & DivideMix* & ELR* & CORES$^2$* & SOP* & 
    UNICON & UNICON$^{+k}$  & DISC & DISC$^{+k}$ \\
     & \cite{li2020dividemix} & \cite{liu2020ELR} & \cite{cheng-instance} & \cite{pmlr-v162-liu22w} & \cite{karim2022unicon} & (ours) & \cite{li2023disc} &  (ours)\\
    \midrule
    69.45 & 74.76 & 72.87 & 73.24 & 73.50 & 74.56 & \textbf{75.13} & 73.30 & 73.87\\
    \bottomrule
  \end{tabular}
  \label{tb:clothing1M}
\end{table}

\noindent\textbf{Results.} Table~\ref{tb:asym} summarizes classification accuracy in asymmetric noise settings, highlighting the advantage of LNL+K in visually similar noise cases. Our adaptation methods consistently outperform the original methods across most noise settings. Notably, FINE$^{+k}$ exhibits significant performance improvement, achieving up to an 8\% increase in accuracy compared to the base FINE method on  CIFAR-100. For CIFAR-100 with a 0.4 noise ratio, the base model SFT achieves 4\% higher accuracy than FINE. However, with the integration of knowledge, FINE$^{+k}$ surpasses the performance of SFT$^{+k}$ by 2\%. These results underscore the importance of investigating LNL+K tasks. The reported gains on Animal-10N~\cite{song2019selfie} in Table~\ref{tb:asym} and Clothing1M~\cite{xiao2015learning} in Table~\ref{tb:clothing1M} further illustrate the advantages of integrating knowledge with LNL+K on general, real-world LNL benchmarks.

\subsection{Discussion}
\label{sec:discussion}
\begin{figure}[t]
\centering
\begin{minipage}[c]{0.5\textwidth}
\includegraphics[width=\textwidth]{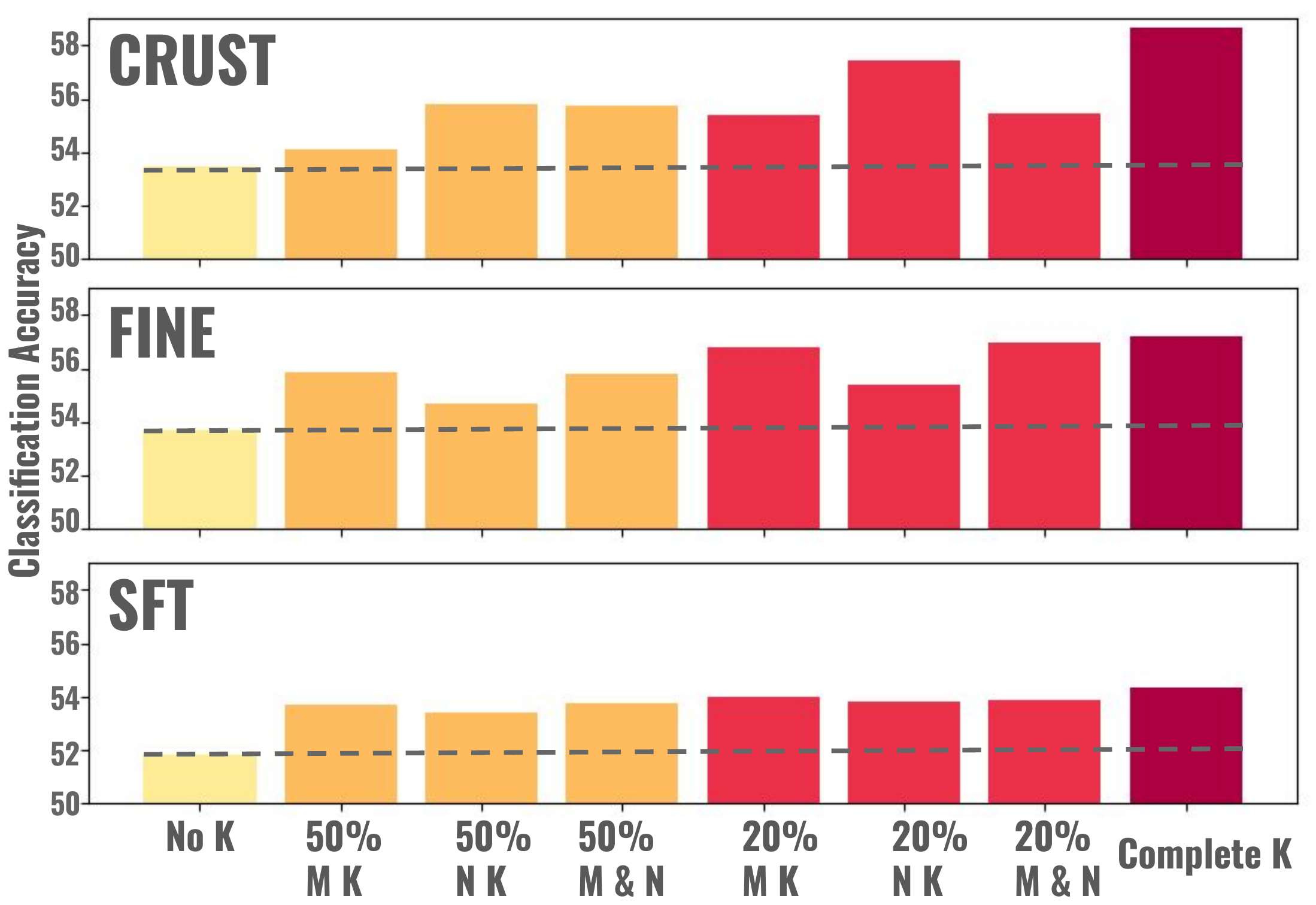}
\end{minipage}%
\hspace{0.5cm}
\begin{minipage}[c]{0.35\textwidth}
\caption{Comparisions of knowledge-adaptive methods with different degrees of noisy noise sources. (MK: missing knowledge, NK: noisy knowledge and M\&N: the combination of these two.) Note: complete knowledge has 50 noise sources. See Section~\ref{sec:discussion} for discussion.}
\label{fig:noisy_source}
\end{minipage}
\end{figure}

\noindent\textbf{Incomplete or noisy knowledge.}
Noise sources need not be strictly complete or clean to provide benefits. \Eg, Fig.~\ref{fig:noisy_source} shows results of missing and incorrect noise sources on CIFAR-100 with a 20\% dominant noise setting. Each recessive class has noise from 50 dominant classes. Missing knowledge (MK): 50\% MK means 25 noise sources are missing. Noisy knowledge (NK): 50\% NK means 25 noise sources are incorrect. M\&N: 50\% M\&N means 25 correct noise sources. We find partial or incomplete knowledge (columns 2-7) is better than no knowledge (column 1) and sometimes only minorly impacts full knowledge (last column).
\smallskip

\noindent\textbf{Learning noise source knowledge from noise transition estimation methods.}
We also explored estimating noise source knowledge using DualT~\cite{yao2020dual} with knowledge-adapted methods. Table \ref{tb:estK} reports performance, where we find combining DualT with our LNL+K methods boosts performance.  Notably, when compared to the original LNL variants from Table~\ref{tb:asym}, our LNL+K models obtain similar or better performance even with estimated noise source knowledge, further validating the importance of our work. 
Noise source knowledge acts as a bridge between noise estimation and detection methods, enabling knowledge-integrated approaches to function even in the absence of prior information.
\smallskip

\begin{table}[t]
\small
\caption{Results of combining noise estimation and noise detection algorithms with knowledge on multiple datasets. We find that using noise estimation with LNL+K can still boost performance. See Section~\ref{sec:discussion} for discussion.}
  \centering
  \setlength{\tabcolsep}{4pt}
  \begin{tabular}{lcccccc}
    \toprule
     & \multicolumn{2}{c}{CIFAR-10~\cite{krizhevsky2009cifar}} & \multicolumn{2}{c}{CIFAR-100~\cite{krizhevsky2009cifar}} & CHAMMI-CP & Animal-10N\\
     \cmidrule(lr){2-3} \cmidrule(lr){4-5} 
     Asym Noise Ratio & \multicolumn{1}{c}{0.2} & \multicolumn{1}{c}{0.4} & \multicolumn{1}{c}{0.2} & \multicolumn{1}{c}{0.4} & \cite{CHAMMI} & \cite{song2019selfie}\\
    \midrule
    
    FINE~\cite{kim2021fine} & 89.07 & 85.51 & 65.42  & 65.11& 67.27 & 81.15\\
    DualT~\cite{yao2020dual}+FINE$^{+k}$ & 89.89 & 88.87 & 66.36  & 62.80& \textbf{70.70} & 81.84\\
    FINE$^{+k}$ & \textbf{90.87} & \textbf{89.15} & \textbf{73.59} & \textbf{72.87} & 67.02 &\textbf{82.27}\\
    \bottomrule
  \end{tabular}
  \label{tb:estK}
\end{table}

\noindent\textbf{Knowledge absorption rate varies for different methods at the same noise settings.}
From the results in Section~\ref{subsec:dominant} and Section~\ref{subsec:asym}, we notice that the accuracy improvements of the adaptation methods vary in different noise settings and methods. We define this different degree of improvement as \textit{knowledge absorption rate}. 
Considering the unified framework of detecting clean labels in Section~\ref{sec:task_definition}, $p(c|x_i)$ and $p(c_n|x_i)$ are important factors to \textit{Knowledge absorption rate}. Our baseline methods represent five different methods of estimating $p(c|x_i)$. The results conclude that noise source knowledge might be more helpful to the feature-based clean sample detection methods in high noise ratios. 
CRUST$^{+k}$ has better performance than FINE$^{+k}$ on high noise ratios in the cell datasets in Table~\ref{tb:dominant}.  A similar observation is made in dominant noise settings, such as the those explored in Table~\ref{tb:sample}, where CRUST$^{+k}$ received a larger boost to performance than other methods, such as FINE$^{+k}$. One possible explanation for this is that $p(c|x_i)$ for FINE$^{+k}$ depends on the category feature distribution while CRUST$^{+k}$ focuses on the feature of a single sample and aims to find the subset with minimum gradient distance sum. In other words, when the noise ratio is high, category feature distribution in FINE$^{+k}$ might be skewed while CRUST$^{+k}$ is less affected by finding the optimal cluster. 
\textit{Knowledge absorption rate} indicates how well an LNL method can transfer to the LNL+K task with noise distribution knowledge, exploring ways to enhance the transferability of LNL methods and optimizing this value are important areas for further investigation.
\smallskip

\section{Conclusion}
This paper introduces a new task, LNL+K, which leverages noise source distribution knowledge when learning with noisy labels. This knowledge is not only beneficial to distinguish clean samples that are ambiguous or out-of-distribution but also necessary when the noise ratio is so high that the noisy samples dominate the class distribution. Instead of comparing the \textit{similarity} of the samples within the same class to detect the clean ones, LNL+K utilizes the \textit{dissimilarity} between the sample and the noise source for detection. We provide a unified framework of clean sample detection for 
LNL+K which we use to adapt state-of-the-art LNL methods, CRUST$^{+k}$, FINE$^{+k}$,  SFT$^{+k}$, UNICON$^{+k}$ and DISC$^{+k}$ to our task.
To create a more realistic simulation of high-noise-ratio settings, we introduce a novel noise setting called \textit{dominant noise}. Results show LNL+K methods have up to 8\% accuracy gains over asymmetric noise and up to 15\% accuracy gains in the dominant noise setting. 
Finally, we discuss the robustness of LNL+K towards incomplete and noisy source knowledge and learning source knowledge from noise estimation methods. We also define \textit{knowledge absorption}, which notes the ranking of LNL methods to our task varies from their LNL performance, indicating that direct investigation of LNL+K is necessary. 
\smallskip

\noindent\textbf{Acknowledgements.} This material is based upon work
supported by the National Science Foundation under Grant No.\ DBI-2134696. Any opinions, findings, and conclusions or recommendations expressed in this material are those of the author(s) and do not necessarily reflect the views of the supporting agencies.



\bibliographystyle{splncs04}
\bibliography{main}







\DeclareRobustCommand\onedot{\futurelet\@let@token\@onedot}
\def\@onedot{\ifx\@let@token.\else.\null\fi\xspace}

\def\eg{\emph{e.g}\onedot} \def\Eg{\emph{E.g}\onedot}
\def\ie{\emph{i.e}\onedot} \def\Ie{\emph{I.e}\onedot}
\def\cf{\emph{cf}\onedot} \def\Cf{\emph{Cf}\onedot}
\def\etc{\emph{etc}\onedot} \def\vs{\emph{vs}\onedot}
\def\wrt{w.r.t\onedot} \def\dof{d.o.f\onedot}
\def\iid{i.i.d\onedot} \def\wolog{w.l.o.g\onedot}
\def\etal{\emph{et al}\onedot}


\title{LNL+K: Enhancing Learning with Noisy Labels Through Noise Source Knowledge Integration Supplementary}
\author{Siqi Wang \and
Bryan A. Plummer }

\authorrunning{S. Wang et al.}
\titlerunning{LNL+K}

\institute{Boston University, Boston MA 02215, USA\\
\email{\{siqiwang,bplum\}@bu.edu}}
\maketitle

\section{LNL+K baseline methods}
\label{sec:app-baselines}
For the convenience of formulating the following equations, recall the notations we defined in Section 3 in the main paper. Dataset $D = \{(x_i, \widetilde{y_i})_{i=1}^n \in R^d \times K\}$, where $K = \{1, 2, ..., k\}$ is the categorical label for $k$ classes. $(x_i, \widetilde{y_i})$ denotes the $i-th$ example in the dataset. $\{\widetilde{y_i}\}_{i=1}^n$ might include noisy labels and the true labels $\{y_i\}_{i=1}^n$ are unknown. Noise transition probability matrix  $P_{k \times k}$, where $P_{ij}$ refers to the probability that a sample in class $i$ is mislabeled as class $j$. A set of label pairs  $LP = \{(i, j) | i, j \in K\}$, where $(i, j)$ indicates that samples in class $i$ are more likely to be mislabeled as class $j$. noise source knowledge $D_{c-ns}$ represents the set of noise source labels of category $c$. \Ie, $D_{c-ns} = \{i | i\in K \land (P_{ic}>0 \lor (i,c) \in LP)\}$. 

\subsection{CRUST$^{+k}$}
The key idea of CRUST~\cite{crust} is from the neural network Jacobian matrix containing all its first-order partial derivatives. It is proved in their work that the neural network has a low-rank Jacobian matrix for clean samples. In other words, data points with clean labels in the same class often have similar gradients clustered closely together. CRUST~\cite{crust} is a feature-based method and this approach can be summarized with settings in Section 3.1. The feature used for selection is the pairwise gradient distance within the class: $g(X_c) = \{d_{x_ix_j}(\mathcal{W}) | x_i, x_j \in X_c\}$, where $d_{x_ix_j}(\mathcal{W}) = \| \nabla L(\mathcal{W}, x_i) - \nabla L(\mathcal{W}, x_j) \|_2$, $\mathcal{W}$ is the network parameters and $L(\mathcal{W}, x_i) = \frac{1}{2} \sum_{x_i \in D} (y_i - f_\theta(\mathcal{W}, x_i))^2$. CRUST~\cite{crust} needs an additional parameter $\beta$ to control the size of the clean selection set $X_c'$. Given $\beta$, the sample $x_i$ is selected as clean if $\|X_c'\| = \beta$ ($\|X_c'\|$ is the size of set $X_c'$) and $x_i \in X_c'$, where $\sum g(X_c')$ has the minimum value. \ie, the selected clean subset $X_c'$ has the most similar gradients clustered together. Thus, we can summarize the similarity metric $M$ for $p(c|x_i)$ as:
\begin{equation}
\begin{aligned}[b]
    M(x_i, \phi_c, \beta)=1 \leftrightarrow \exists X_c'\subset X_c \land  \|X_c'\|= \beta,\\
    s.t. \: x_i \in X_c' \land (\forall \|X_c''\|= \beta \land X_c'' \subset X_c, \\
    \sum g(X_c') \leq \sum g(X_c'')),
\label{eq:crust-m}
\end{aligned}
\end{equation}
otherwise $M(x_i, \phi_c, \beta)=0$.
Thus, we can get the propositional logic of CRUST:
\begin{equation}
y_i =c \leftrightarrow \widetilde{y_i} = c \land p(c|x_i) =1 \leftrightarrow M(x_i, \phi_{\widetilde{y_i}}, \beta)=1. 
\label{eq:crust-p}
\end{equation}
To adapt CRUST to CRUST$^{+k}$ with noise source distribution knowledge. 
from Eq.2 in the main paper, we have
\begin{equation}
\begin{aligned}[b]
\widetilde{y_i} = c \land y_i \neq c \leftrightarrow p(c|x_i) \leq \max(\{p(c_n|x_i) | c_n \in D_{c-ns}\})\\ 
\leftrightarrow \exists c_n \in D_{c-ns}\: s.t. \:  p(c_n|x_i) \geq p(c|x_i) \\
\leftrightarrow \exists c_n \in D_{c-ns}\: s.t. \:  p(c_n|x_i) = 1. 
\end{aligned}
\label{eq:crust-rewrite-eq2}
\end{equation}

To get $p(c_n|x_i)$, we first mix $x_i$ with all the samples in $X_{c_n}$, \ie, $X_{c_n+} = \{x_i\} \cup X_{c_n}$. Then apply CRUST on this mix set, \ie, calculate the loss towards label $c_n$ and select the clean subset $X_{c_n+}'$. if $x_i \in X_{c_n+}'$, then $p(c_n|x_i) = 1$. 
Here is the formulation of CRUST$^{+k}$, we modify $L(\mathcal{W}, x_i)$ to $L(\mathcal{W}, x_i, c) = \frac{1}{2} \sum_{x_i \in D} (c - f_\theta(\mathcal{W}, x_i))^2$, where we calculate the loss to any certain categories, not limited to the loss towards the label. Similarly, we have  $d_{x_ix_j}(\mathbf{W}, c) = \| \nabla L(\mathcal{W}, x_i, c) - \nabla L(\mathcal{W}, x_j, c) \|_2$,  $g(X_{c_n+}, c_n) = \{d_{x_ix_j}(\mathbf{W}, c_n) | x_i, x_j \in X_{c_n+}\}$. We use $\gamma$ to represent the subset size of $X_{c+c_n}$, which is decided by $\beta$ and noise source distribution. Finally, we get the similarity metric $M(x_i, \phi_{c_n+}, \gamma)$ as:
\begin{equation}
\begin{aligned}[b]
    M(x_i, \phi_{c_n+}, \gamma) = 1 \leftrightarrow \exists X_{c_n+}'\subset X_{c_n+} \land \|X_{c_n+}'\|= \gamma,\\
    s.t. \:  x_i \in X_{c_n+}' \land (\forall \|X_{c_n+}''\|= \gamma \land X_{c_n+}'' \subset X_{c_n+}, \\
    \sum g(X_{c_n+}', c_n) \leq \sum g(X_{c_n+}'', c_n)),
\end{aligned}
\label{eq:CRUST$^{+k}$-m}
\end{equation}otherwise $M(x_i, \phi_{c_n+}, \gamma)=0$. Combining Eq.2 in the main paper, Eq.\ref{eq:crust-p}, and Eq.\ref{eq:CRUST$^{+k}$-m},  $p(c|x_i)$ of CRUST$^{+k}$ method is:
\begin{equation}
\begin{aligned}[b]
y_i = c 
\leftrightarrow \widetilde{y_i} = c \land (\forall c_n \in D_{c-ns},  p(c_n|x_i) < p(c|x_i)) \\
\leftrightarrow \widetilde{y_i} = c \land (\forall c_n \in D_{c-ns},  p(c_n|x_i) = 0) \\
\leftrightarrow \widetilde{y_i} = c \land (\forall c_n \in D_{c-ns},  M(x_i, \phi_{c_n+}, \gamma) = 0).
\end{aligned}
\end{equation}

\subsection{FINE$^{+k}$}
Filtering Noisy instances via their Eigenvectors(\textit{FINE})~\cite{kim2021fine} selects clean samples with the feature-based method. Let $f_{\theta^*}(x_i)$ be the feature extractor output and $\Sigma_c$ be the gram matrix of all features labeled as category $c$. The alignment is defined as the cosine distance between feature $\overrightarrow{f_{\theta^*}(x_i)}$ and $ \overrightarrow{c}$, which is the eigenvector of the $\Sigma_c$ and can be treated as the feature representation of category $c$. FINE fits a Gaussian Mixture Model (GMM) on the alignment distribution to divide samples to clean and noisy groups - the clean group has a larger mean value, which refers to a better alignment with the category feature representation. In summary, feature mapping function $g(x_i, c) = <\overrightarrow{f_{\theta^*}(x_i)}, \overrightarrow{c}>$, and mixture of Gaussian distributions $\phi_c = \mathcal{N}_{clean} + \mathcal{N}_{noisy} = \mathcal{N}(\mu_{g(X_{c-clean})}, \sigma_{g(X_{c-clean})}) + \mathcal{N}(\mu_{g(X_{c-noisy})}, \sigma_{g(X_{c-noisy})})$, where $\mu_{g(X_{c-clean})} > \mu_{g(X_{c-noisy})}$. The similarity metric
\begin{equation*}
M(x_i, \phi_c)=
\left\{
\begin{array}
{r@{\quad:\quad}l}
1 & \mathcal{N}_{clean}(g(x_i,c)) > \mathcal{N}_{noisy}(g(x_i,c)) \\
0 & \mathcal{N}_{clean}(g(x_i,c)) \leq \mathcal{N}_{noisy}(g(x_i,c)).
\end{array}
\right.
\end{equation*}

Thus, we have
\begin{equation}
y_i  =c \leftrightarrow \widetilde{y_i} = c \land p(c|x_i)=1  \leftrightarrow M(x_i, \phi_{\widetilde{y_i}})= 1.
\end{equation}
Next, we show our design of FINE$^{+k}$ with noise source distribution knowledge. The key difference between FINE and FINE$^{+k}$ is that we use the alignment score of the noise source class. 
For a formal description of FINE$^{+k}$, We define $g_k(x_i, c, c_n) = g(x_i, c) - g(x_i, c_n)$. Similar to FINE, FINE$^{+k}$ fits a GMM on $g_k(X_c, c, c_n)$, so we have $g_k(X_c, c, c_n) \sim \phi_{k-\{c+c_n\}} = \mathcal{N}_{close-c} + \mathcal{N}_{close-c_n} $, where $\mu_{close-c} > \mu_{close-c_n}$. This can be interpreted in the following way: Samples aligning better with category $c$ should have larger $g(x_i, c)$ values and smaller $g(x_i, c_n)$ values according to the assumption, thus the greater the $g_k(x_i, c, c_n)$, the closer to category $c$, vice versa, the smaller the $g_k(x_i, c, c_n)$, the closer to category $c_n$.
Then we have
\begin{equation}
    \begin{aligned}[b]
    M(x_i, \phi_{k-\{c+c_n\}})= 1 \\
    \leftrightarrow \mathcal{N}_{close-c}(g_k(x_i, c, c_n)) > \mathcal{N}_{close-c_n}(g_k(x_i, c, c_n))
    \end{aligned}
\end{equation}
otherwise $M(x_i, \phi_{k-\{c+c_n\}})=0$.
By combining with Eq.2 in the main paper, we have
\begin{equation}
\begin{split}
y_i  =c \leftrightarrow \widetilde{y_i}= c \land (\forall c_n \in D_{c-ns}, p(c|x_i) > p(c_n|x_i))\\
\leftrightarrow y_i= c \land (\forall c_n \in D_{c-ns}, M(x_i, \phi_{k-\{c+c_n\}})= 1 ).
\end{split}
\label{eq:5}
\end{equation}

\subsection{SFT$^{+k}$}
SFT~\cite{wei2022self} detects noisy samples according to predictions stored in a memory bank $\mathcal{M}$. $\mathcal{M}$ contains the last $T$ epochs' predictions of each sample. A sample $x_i$ is detected as noisy if a fluctuation event occurs, \ie, the sample classified correctly at the previous epoch $t_1$ is misclassified at $t_2$, where $t_1 < t_2$. The occurrence of the fluctuation event can be formulated as $fluctuation(x_i, y_i)=1$, otherwise $fluctuation(x_i, y_i)=0$ \ie, 
\begin{equation}
\begin{aligned}[b]
    \text{fluctuation}(x_i, y_i)=1 \\
    \leftrightarrow \exists t_1, t_2  \in \{t-T, \cdots ,T\} \land t_1 < t_2 \\
    s.t.  \:  f_\theta(x_i)^{t_1} = \widetilde{y_i} \land f_\theta(x_i)^{t_2} \neq \widetilde{y_i},
\end{aligned}
\end{equation}
where $f_\theta(x_i)^{t_1}$ represents the prediction of $x_i$ at epoch $t_1$. SFT is a probability-distribution-based approach and can fit our probabilistic model as follows. The propositional logic of SFT is,
\begin{equation}
p(c|x_i)=
\left\{
\begin{array}
{r@{\quad:\quad}l}
1 & \widetilde{y_i} = c \land \text{fluctuation}(x_i, \widetilde{y_i}) = 0\\
0 & otherwise.
\end{array}
\right.
\label{eq:sft-p}
\end{equation}
\Ie, SFT$^{+k}$ applies the noise source distribution knowledge to SFT by stricting the constraints of fluctuation. The fluctuation events only occur when the previous correct prediction is misclassified as the noise source label. 
Thus, we define SFT$^{+k}$ fluctuation as,
\begin{equation}
\begin{aligned}[b]
    \text{fluctuation}(x_i, y_i, D_{y_i-ns})=1 \\
    \leftrightarrow \exists c_n \in D_{y_i-ns}, \exists t_1, t_2  \in \{t-T, \cdots ,T\} \land t_1 < t_2, \\
    s.t.  \:  f_\theta(x_i)^{t_1} = y_i \land f_\theta(x_i)^{t_2} = c_n.
\end{aligned}
\label{eq:sft-new-fluctuation}
\end{equation}
Combining Eq.2 in the main paper, Eq.~\ref{eq:sft-p} and Eq.~\ref{eq:sft-new-fluctuation},  SFT$^{+k}$ detects $x_i$ with clean label $y_i=\widetilde{y_i}=c$ with $p(c|x_i)$ as:
\begin{equation}
\begin{aligned}[b]
y_i=c \\
\leftrightarrow \widetilde{y_i} = c \land p(c|x_i) > \max(\{p(c_n|x_i) | c_n \in D_{c-ns}\})\\
\leftrightarrow \widetilde{y_i} = c \land p(c|x_i) = 1 \\
\leftrightarrow \widetilde{y_i} = c \land fluctuation(x_i, \widetilde{y_i}, D_{\widetilde{y_i}-ns}) = 0.
\end{aligned}
\end{equation}

\subsection{UNICON$^{+k}$}
UNICON~\cite{karim2022unicon} estimate the clean probability by using Jensen-Shannon divergence (JSD) $d_i$, which is a measure of distribution disagreement. JSD is defined by KLD, which is the Kullback-Leibler divergence function. We follow the same JSD definition as UNICON in the adaptation method. Given the predicted probability $p_i$ and label $\widetilde{y_i}$, $d_i = JSD(\widetilde{y_i}, p_i)$. The value of $d_i$ ranges from 0 to 1 and the smaller the $d_i$ is, the higher the probability of $\widetilde{y_i}$ being clean. A cutoff value $d_{cutoff}$ is used to select clean samples. To summarize, the propositional logic of UNICON is,
\begin{equation}
    \begin{aligned}[b]
   p(c|x_i) = 1-JSD(x_i, y_i) \\
    \leftrightarrow  y_i = c \land JSD(x_i, y_i) < d_{cutoff}
    \end{aligned}
    \label{eq:unicon-p}
\end{equation}
otherwise $p(c|x_i)=0$.
Then noise source knowledge is integrated with our unified framework:
\begin{equation}
\begin{split}
y_i=c \leftrightarrow \widetilde{y_i} = c \land p(c|x_i) > \max(\{p(c_n|x_i) | c_n \in D_{c-ns}\})\\
\leftrightarrow \widetilde{y_i} = c \land (\forall c_n \in D_{c-ns}, JSD(x_i,\widetilde{y_i})<JSD(x_i,c_n)) .
\end{split}
\end{equation}

\subsection{DISC$^{+k}$}
DISC~\cite{li2023disc} employs weak and strong augmentations on each single noisy labeled data and divides samples into \textit{Clean}, \textit{Hard}, and \textit{Purified} sets according to the prediction confidences on the two-augmentation views. The clean set is determined with the prediction confidence in weak view $Conf_w$, confidence in strong view $Conf_s$, and dynamic instance specific thresholds (\textit{DIST}) for weak and strong views $\tau_w$ and $\tau_s$.
The \textit{DIST} is defined as,
\begin{equation}
    \begin{aligned}[b]
    \tau(x,t) = \lambda \tau(t-1) + (1-\lambda) \max(\{Conf(c, x) | c \in K\} , \tau(0) = 0.
    \end{aligned}
    \label{eq:disc-dist}
\end{equation}
where $Conf(c, x)$ represents the prediction confidence in sample $x$ for class $c$. Then the clean set selected at time $t$ of DISC can be defined as,
\begin{equation}
    \begin{aligned}[b]
        C (t) = \{x_i | Conf_w(\widetilde{y_i}, x_i) > \tau_w(x_i, t)\} \cap  \{x_i | Conf_s(\widetilde{y_i}, x_i) > \tau_s(x_i, t)\}.
    \end{aligned}
    \label{eq:disc-C}
\end{equation}
According to the Eq.1 in the main paper, sample $x_i$ is clean at time $t$ in DISC is detected with,
\begin{equation}
    \begin{aligned}[b]
     y_i=c \leftrightarrow \widetilde{y_i} = c \land ( Conf_w(\widetilde{y_i}, x_i) > \tau_w(x_i, t) \land Conf_s(\widetilde{y_i}, x_i) > \tau_s(x_i, t)).
    \end{aligned}
    \label{eq:disc}
\end{equation}
DISC$^{+k}$ introduces class-wise comparisons and makes adaptations to the instance-specific threshold. Instead of taking the maximum confidence from all the class (\ie $\max(\{Conf(c, x) | c \in K\}$), DISC$^{+k}$ only selects maximum from the labeled class and noise source classes. To be specific, we have,
\begin{equation}
    \begin{aligned}[b]
    \tau^{+k}(x,\widetilde{y},t) = \lambda \tau(t-1) + (1-\lambda) \max(\{Conf(c, x) | c \in D_{\widetilde{y}-ns}\} \cup \{Conf(\widetilde{y}, x)\}).
    \end{aligned}
    \label{eq:discK-dist}
\end{equation}
According to the Eq.2 in the main paper, the noise source knowledge is integrated as,
\begin{equation}
    \begin{aligned}[b]
        y_i=c \leftrightarrow \widetilde{y_i} = c \land ( Conf_w(\widetilde{y_i}, x_i) > \tau_w^{+k}(x_i,\widetilde{y_i},t) \land Conf_s(\widetilde{y_i}, x_i) > \tau_s^{+k}(x_i,\widetilde{y_i},t)).
    \end{aligned}
    \label{eq:discK}
\end{equation}

\section{Datasets}
\subsection{CIFAR datasets~\cite{krizhevsky2009cifar} with synthesized noise}

\textbf{Dominant noise}
\label{sec:app-dataset-ctl}
There are \textit{recessive} and \textit{dominant} classes in dominant noise. For CIFAR-10~\cite{krizhevsky2009cifar}, category index of the last 5 are \textit{recessive} classes and the first five are \textit{dominant} classes. In other words, category index 6-10 samples might be mislabeled as label index 1-5. Different numbers of samples are mixed for different noise ratios so that the dataset is still balanced after mislabeling. Table \ref{tb:dominant-noise} shows the number of samples per category for each noise ratio. Notably, the dataset is balanced after the mislabeling. In each recessive class, there are multiple noise sources, with all dominant classes serving as the noise sources. To illustrate, in CIFAR-10~\cite{krizhevsky2009cifar}, classes 6-10 are considered recessive, and instances of these classes might be incorrectly labeled as dominant classes 1-5.  To maintain balance after mislabeling, we adopt an unbalanced sampling approach to construct the dataset. For instance, with a noise ratio of 0.5 in the CIFAR-10 dataset, we sample 1250 instances for each \textit{dominant} class and 3750 instances for each \textit{recessive} class. After mislabeling 1250 samples to the \textit{dominant} class for each \textit{recessive} class, there are 2500 samples in each class.

\noindent\textbf{Asymmetric noise.}
\label{sec:app-dataset-asym}
Labels are corrupted to visually similar classes. Pair $(C_1, C_2)$ represents the samples in class $C_1$ and $C_2$ are possibly mislabeled as each other. Noise ratios in the experiments are only the noise ratio in class pairs, \ie not the overall noise ratio. Here are the class pairs of CIFAR-10 and CIFAR-100~\cite{krizhevsky2009cifar} for asymmetric noise. 
\textbf{CIFAR-10}~\cite{krizhevsky2009cifar} (trucks, automobiles), (cat, dog), (horse, deer).
\textbf{CIFAR-100}~\cite{krizhevsky2009cifar} (beaver, otter), (aquarium fish, flatfish), (poppies, roses),
(bottles, cans),
(apples, pears),
(chair, couch),
(bee, beetle),
(lion, tiger),
(crab, spider),
(rabbit, squirrel),
(maple, oak),
(bicycle, motorcycle).

\begin{table}[tbh]
    \centering
    \caption{Sample composition for CIFAR-10/CIFAR-100~\cite{krizhevsky2009cifar} dominant noise.}
    \begin{tabular}{lccc}
        \toprule
        \multicolumn{4}{l}{CIFAR-10~\cite{krizhevsky2009cifar} Dominant Noise}\\
        \hline
        Noise ratio & 0.2 & 0.5 & 0.8\\
        Dominant class & 2000 & 1250 & 500\\
        Recessive class & 3000 & 3750 & 4500\\
        \hline
        \multicolumn{4}{l}{CIFAR-100~\cite{krizhevsky2009cifar} Dominant Noise}\\
        \hline
        Noise ratio & 0.2 & 0.5 & 0.8\\
        Dominant class & 200 & 125 & 50\\
        Recessive class & 300 & 375 & 450\\
    \bottomrule
    \end{tabular}
  \label{tb:dominant-noise}
\end{table}

\subsection{Cell dataset BBBC036~\cite{bray2016cell}}
\label{sec:app-dataset-BBBC036}
For our experiments, we subsampled 100 treatments to evaluate natural noise. Table \ref{tb:treatments} shows the treatment list. ("NA" refers to the \textit{control} group, \ie no treatment group.)

\begin{table*}[tbh]
  \centering
  \caption{Treatments used from the BBBC036 dataset~\cite{bray2016cell}}
  \begin{tabular}{lllll}
  \toprule
    NA & BRD-K88090157 & BRD-K38436528 & BRD-K07691486 & BRD-K97530723\\
    BRD-A32505112 & BRD-K21853356 & BRD-K96809896 & BRD-A82590476 & BRD-A95939040 \\
    BRD-A53952395 & BRD-A64125466 & BRD-A99177642 & BRD-K90574421 & BRD-K07507905 \\
BRD-K62221994 & BRD-K62810658 & BRD-K47150025 & BRD-K17705806 & BRD-K85015012 \\
BRD-K37865504 & BRD-A52660433 & BRD-K66898851 & BRD-K15025317 & BRD-K37392901 \\
BRD-K91370081 & BRD-K39484304 & BRD-K03842655 & BRD-K76840893 & BRD-K62289640 \\
BRD-K14618467 & BRD-K52313696 & BRD-K43744935 & BRD-K86727142 & BRD-K21680192 \\
BRD-K06426971 & BRD-K24132293 & BRD-K68143200 & BRD-K08554278 & BRD-K78122587 \\
BRD-A47513740 & BRD-K18619710 & BRD-A67552019 & BRD-K17140735 & BRD-K30867024 \\
BRD-K36007650 & BRD-K51318897 & BRD-K90382497 & BRD-K00259736 & BRD-K95435023 \\
BRD-K52075040 & BRD-K03642198 & BRD-K47278471 & BRD-K17896185 & BRD-K95603879 \\
BRD-A70649075 & BRD-K02407574 & BRD-A90462498 & BRD-K67860401 & BRD-A64485570 \\
BRD-K88429204 & BRD-A49046702 & BRD-K50841342 & BRD-K35960502 & BRD-K77171813 \\
BRD-K54095730 & BRD-K93754473 & BRD-K22134346 & BRD-K72703948 & BRD-K31342827 \\
BRD-K31542390 & BRD-K18250272 & BRD-K00141480 & BRD-K37991163 & BRD-K13533483 \\
BRD-K67439147 & BRD-A91008255 & BRD-K39187410 & BRD-K26997899 & BRD-K89732114 \\
BRD-K50135270 & BRD-K95237249 & BRD-K44849676 & BRD-K20742498 & BRD-K31912990 \\
BRD-K96799727 & BRD-K09255212 & BRD-A89947015 & BRD-K78364995 & BRD-K49294207 \\
BRD-K08316444 & BRD-K89930444 & BRD-K50398167 & BRD-K47936004 & BRD-A72711497 \\
BRD-A97104540 & BRD-A50737080 & BRD-K80970344 & BRD-K50464341 & BRD-K97399794 \\
    \bottomrule
    \end{tabular}
  \label{tb:treatments}
\end{table*}

\subsection{Cell dataset CHAMMI-CP~\cite{CHAMMI}}
\label{sec:app-dataset-CP}
Three compounds with a \textit{control} group are selected for our experiments: BRD-A29260609 (weak reaction), BRD-K04185004 (medium reaction), and BRD-K21680192 (strong reaction).

\subsection{Clothing1M dataset~\cite{xiao2015learning}}
We conducted experiments on the Clothing1M dataset~\cite{clothing1m}, the noise source knowledge is summarized according to the confusion matrix from the dataset~\cite{clothing1m}. We use $ a \to b$ to represent $a$ as the noise source of $b$. The prior noise knowledge is: \textit{Chiffon} $\to$ \textit{Shirt}, \textit{Sweater} $\to$ \textit{Knitwear}, \textit{Knitwear} $\to$ \textit{Sweater}, \textit{Jacket} $\to$ \textit{Windbreaker}, \textit{Windbreaker} $\to$ \textit{Down coat}, and \textit{Vest}$\to$ \textit{Dress}.

\section{Feature extractors for each dataset}
We used a pre-trained ResNet34~\cite{he2016deep} on CIFAR-10/CIFAR-100~\cite{krizhevsky2009cifar} for all approaches (UNICON~\cite{karim2022unicon} trains on two networks), ResNet50~\cite{he2016deep} on Animal-10N~\cite{song2019selfie} and Clothing1M~\cite{xiao2015learning} datasets.
For experiments on BBBC036~\cite{bray2016cell} we used an Efficient B0~\cite{tan2019efficientnet} for all methods and all methods used ConvNet~\cite{liu2022convnet} for CHAMMI-CP~\cite{CHAMMI} dataset. To support the 5 channel images in cell datasets, we replaced the first convolutional layer in the network to support the new image dimensions.

\section{Hyperparameters}
For a fair comparison, we use the same hyperparameter settings as in prior work~\cite{crust, kim2021fine, wei2022self, karim2022unicon, li2023disc} for CIFAR-10/CIFAR-100~\cite{krizhevsky2009cifar} datasets. Hyperparameters of the cell dataset BBBC036~\cite{bray2016cell} were set via grid search using the validation set. All the experiments use the same batch size of 128. "fl-ratio" of CRUST~\cite{crust} and CRUST$^{+k}$, which controls the size of selected clean samples is set as the same as the noise ratio in synthesized noise and set as 0.6 in cell dataset BBBC036~\cite{bray2016cell} and CHAMMI-CP~\cite{CHAMMI}, 0.9 in Animal10N~\cite{song2019selfie} and Clothing1M~\cite{xiao2015learning}. All the other hyperparameters for each dataset are summarized in Table \ref{tab:hyperparam}. 

\begin{table}
    \centering
    \caption{Hyperparameters for each dataset.}
    \begin{tabular}{l|c|c|c}
    \toprule
      & learning rate & warm-up epochs & total number of epochs\\
      \hline
     CIFAR-10/CIFAR-100~\cite{krizhevsky2009cifar} & 1e-2 & 40 & 120\\
     BBBC036~\cite{bray2016cell} & 2e-4 & 10 & 100\\
     CHAMMI-CP~\cite{CHAMMI} & 2e-4 & 5 & 30\\
     Animal10N~\cite{song2019selfie} & 5e-3 & 3 & 30\\
     Clothing1M~\cite{xiao2015learning} & 5e-2 & 0 & 200\\
     \bottomrule
    \end{tabular}
    \label{tab:hyperparam}
\end{table}

\section{Additional results on lower noise ratio of dominant noise}
We also performed experiments with 0.2 dominant noise on CIFAR-10/CIFAR-100~\cite{krizhevsky2009cifar} datasets. The results in Table~\ref{tb:0.2-dominant} demonstrate that knowledge integration is also beneficial in cases of lower noise ratios, showcasing the broad applicability of LNL+K across a range of noise levels from 0.2 to 0.8.

\begin{table}[t]
  \caption{0.2 Dominant noise results on CIFAR-10 and CIFAR-100 dataset. The best test accuracy is marked in bold, and the better result between LNL and LNL+K methods is marked with underlined. We find incorporating source knowledge helps in almost all cases.}
  \centering
    \begin{tabular}{lcc}
    \toprule
     & {CIFAR-10~\cite{krizhevsky2009cifar}} & {CIFAR-100~\cite{krizhevsky2009cifar}} \\
    \midrule
    Baseline & 85.47\scriptsize{$\pm$0.52}  & 50.37\scriptsize{$\pm$0.45} \\
    DualT~\cite{yao2020dual} & 86.55\scriptsize{$\pm$0.06}  & 34.88\scriptsize{$\pm$0.11} \\
    GT-T & 88.09\scriptsize{$\pm$0.04}  & 59.32\scriptsize{$\pm$0.14}\\ 
    SOP~\cite{pmlr-v162-liu22w} & 89.86\scriptsize{$\pm$0.40}  & 62.47\scriptsize{$\pm$0.47} \\
    \midrule
    CRUST~\cite{crust} & 88.21\scriptsize{$\pm$0.22} & 53.48\scriptsize{$\pm$0.80} \\
    CRUST$^{+k}$ & \underline{89.53\scriptsize{$\pm$0.05}} & \underline{58.69\scriptsize{$\pm$0.50}} \\
    \midrule
    FINE~\cite{kim2021fine} & 86.23\scriptsize{$\pm$0.30} & 53.68\scriptsize{$\pm$1.54} \\
    FINE$^{+k}$ & \underline{88.69\scriptsize{$\pm$0.06}} & \underline{57.22\scriptsize{$\pm$1.16}} \\
    \midrule
    SFT~\cite{wei2022self} & 89.48\scriptsize{$\pm$0.21} & 51.82\scriptsize{$\pm$0.67} \\
    SFT$^{+k}$ & \underline{89.78\scriptsize{$\pm$0.03}} & \underline{54.36\scriptsize{$\pm$0.48}}\\
    \midrule
    UNICON~\cite{karim2022unicon} & 90.82\scriptsize{$\pm$0.14} & 63.28\scriptsize{$\pm$0.32} \\
    UNICON$^{+k}$ & \underline{90.83\scriptsize{$\pm$0.11}} & \underline{66.77\scriptsize{$\pm$0.54}} \\
    \midrule
    DISC~\cite{li2023disc} & 93.10\scriptsize{$\pm0.12$} & 69.75\scriptsize{$\pm0.13$}\\
    DISC$^{+k}$ & \underline{\textbf{93.55}\scriptsize{$\pm$0.03}} & \underline{\textbf{70.02}\scriptsize{$\pm$0.30}}\\
    \bottomrule
  \end{tabular}

  \label{tb:0.2-dominant}
\end{table}

\section{Ethical considerations}
This study was conducted with biological images of human bone osteosarcoma cells, an immortalized cell line used for research purposes only. The images or data in this study do not contain patient information of any kind. The use of these images, and the algorithms to analyze them, is to test the effects of treatments. Automating drug discovery has positive impacts on society, specifically the potential to help find cures for diseases of pressing need around the world in shorter times, and utilizing fewer resources. The proposed methods could be used to optimize drugs that harm people; we do not intend that as an application, and we expect regulations in biological labs to prevent such uses.


\end{document}